\pdfoutput=1
\documentclass[11pt]{article}

\usepackage[final]{acl}

\usepackage{times}
\usepackage{latexsym}
\usepackage{amsmath}

\usepackage{subcaption}
\usepackage{graphicx}

\usepackage[T1]{fontenc}
\usepackage[utf8]{inputenc}

\usepackage{microtype}

\usepackage{inconsolata}
\usepackage{graphicx}
\usepackage{xspace}
\RequirePackage{todonotes}
\usepackage{multirow} % for table formatting
\usepackage{booktabs} % for table formatting
\usepackage{multirow} % for table formatting
\usepackage{float}  % 使用[H]

\title{Enhancing Semantic Consistency of Large Language Models through Model Editing: An Interpretability-Oriented Approach}

\newcommand*{\samethanks}[1][\value{footnote}]{\footnotemark[#1]}

\author{Jingyuan Yang\textsuperscript{1,2}, Dapeng Chen\textsuperscript{2}, Yajing Sun\textsuperscript{2}, Rongjun Li\textsuperscript{2}, Zhiyong Feng\textsuperscript{1,}\thanks{Corresponding Author} \and Wei Peng\textsuperscript{2,}\samethanks  \\
\textsuperscript{1}College of Intelligence and Computing, Tianjin University \\
\textsuperscript{2}IT Innovation and Research Center, Huawei Technologies \\
\{yangjingyuan2, chendapeng8, sunyajing4, lirongjun3, peng.wei1\}@huawei.com\\ zyfeng@tju.edu.cn}

\begin{document}
\maketitle
\begin{abstract}
A Large Language Model (LLM) tends to generate inconsistent and sometimes contradictory outputs when presented with a prompt that has equivalent semantics but is expressed differently from the original prompt. To achieve semantic consistency of an LLM, one of the key approaches is to finetune the model with prompt-output pairs with semantically equivalent meanings. Despite its effectiveness, a data-driven finetuning method incurs substantial computation costs in data preparation and model optimization. In this regime, an LLM is treated as a ``black box'', restricting our ability to gain deeper insights into its internal mechanism. In this paper, we are motivated to enhance the semantic consistency of LLMs through a more interpretable method (i.e., model editing) to this end. We first identify the model components (i.e., attention heads) that have a key impact on the semantic consistency of an LLM. We subsequently inject biases into the output of these model components along the semantic-consistency activation direction. It is noteworthy that these modifications are cost-effective, without reliance on mass manipulations of the original model parameters. Through comprehensive experiments on the constructed NLU and open-source NLG datasets, our method demonstrates significant improvements in the semantic consistency and task performance of LLMs. Additionally, our method exhibits promising generalization capabilities by performing well on tasks beyond the primary tasks.
\end{abstract}

% \authornote
\section{Introduction}
\begin{figure}[h]
    \centering
    \includegraphics[width=1.0 \columnwidth]{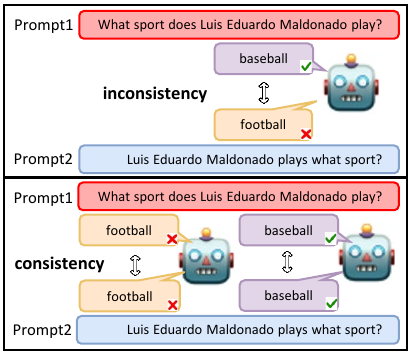}
    \caption{Inconsistency arises when prompts sharing equivalent semantics produce different outcomes, while consistency is achieved when their outputs remain consistently identical, irrespective of their accuracy.}
    \label{fig:my_image1}
\end{figure}

The field of Natural Language Processing (NLP) is experiencing a paradigm shift with the advent of Large Language Models (LLMs). These models have demonstrated remarkable capabilities in various tasks such as sentiment classification~\citep{wang2023chatgpt}, machine translation~\citep{hendy2023good}, and summarization~\citep{pu2023summarization}. However, an LLM tends to generate inconsistent and sometimes contradictory outputs when presented with a prompt that has equivalent semantics but is expressed differently from the original prompt. Such behavior is referred to as the issue of ``semantic consistency'' \citep{gan2023sensitivity, rabinovich2023predicting, raj2022measuring}, largely limiting the application of LLMs to real-world scenarios. For specific instances of inconsistency and consistency, please refer to Figure \ref{fig:my_image1}.    

Current mainstream solutions involve prompt engineering or data-driven methods to handle the problem of semantic consistency. For example, \citet{raj2023semantic} proposed a prompt strategy called ‘Ask-to-Choose’ (A2C) to improve the semantic consistency of LLMs, but this method requires carefully designed prompts. Applying a data-driven supervised fine-tuning method  (SFT) ~\citep{ouyang2022training} to finetune an LLM with prompt-output pairs with semantically equivalent meanings is another effective approach. Despite their effectiveness, these methods incur substantial computation costs in data preparation and model optimization. Furthermore, these methods treat an LLM as a ``black box'', restricting our ability to gain deeper insights into its underlying causes of the semantic consistency problem. 

% Our first step is to create semantically consistent prompt pairs. Our approach 

To address the limitations of previous methods to enhance the semantic consistency of LLMs, we propose a method based on model editing that can locate the internal model components (i.e., attention heads) responsible for generating semantic inconsistency. We subsequently inject biases into the outputs of these model components along semantic consistency activation directions. 
This strategy aims to shift the outputs of the key model components toward a direction resilient to variations of synonymous prompts.

In order to comprehensively evaluate our proposed method under varying prompts, we have constructed relevant NLU-task datasets in addition to utilizing existing evaluation datasets for NLG-task. We leverage the paraphrasing capability of GPT-4\footnote{\url{https://platform.openai.com/docs/api-reference/chat}} to construct the RobustSST2, RobustMRPC, and RobustBOOLQ datasets. These datasets cover a wide range of tasks, including the sentiment classification dataset SST2 \citep{socher2013recursive}, the text similarity dataset MRPC \citep{dolan2005automatically}, and the question-answering dataset BOOLQ \citep{clark2019boolq}. 

Our method has shown significant enhancements in both semantic consistency and task performance on publicly available NLG datasets and our constructed NLU datasets. Furthermore, our method also achieve positive results in out-of-domain experiments, demonstrating a solid generalization capability. In summary, our contributions are two-fold:

\begin{itemize}
    \item To the best of our knowledge, we are the first to use a model editing approach to address the issue of prompt semantic inconsistency. Through this interpretability-oriented method, we can precisely diagnose the internal components contributing to semantic consistency. By directly injecting biases into the model, our method avoids mass-manipulating model parameters, resulting in a significant saving in GPU hour (up to 23 times faster, shown in Table \ref{tab:gpu_hour}) in a typical task compared to a traditional SFT approach. 
   
    \item We have curated three datasets, designed to address the absence of NLU semantic consistency evaluation benchmark. The datasets will be released to the community to foster research along this line. 
\end{itemize}

\section{Related Work}
\begin{figure*}[tbp]
    \centering
    \includegraphics[width=1\textwidth, height=0.28\textwidth]{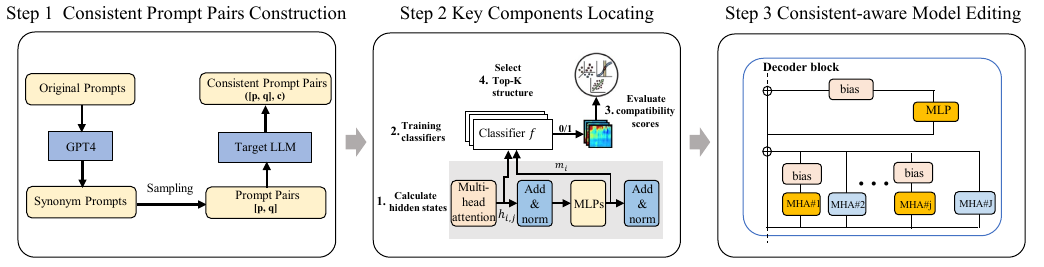}
    \caption{The flowchart of our method. Our method has three main steps: (1) We first construct the prompt pairs $[p, q]$ with consistency evaluation label $c$. (2) Based on these pairs, we perform key-components locating, which selects the top-K (accuracy) components by training and evaluating classifiers based on these components' output hidden states and related consistency evaluation labels. If a classifier has high accuracy, the component and LLM will behave very similarly (compatible), which suggests that the component is highly likely to be responsible for the inconsistency errors, as mentioned previously. (3) For the selected top-K components, we add biases to the hidden states of these components,  which will shift the original activations of these components toward more semantically consistent directions.}
    \label{fig:my_image}
\end{figure*}

\noindent \textbf{Semantic Consistency.}
The study of semantic consistency originated from investigations into Masked Language Models (MLMs) like BERT and Roberta. \citet{elazar-etal-2021-measuring} revealed significant semantic inconsistency in the factual information extracted from these MLMs when subjected to paraphrasing. Building on this, \citet{fierro-sogaard-2022-factual} extended the examination of semantic consistency to a multilingual context, disclosing that inconsistency issues are not confined to English but are prevalent across various other languages. Despite the significant shift in the Natural Language Processing (NLP) paradigm instigated by Large Language Models (LLMs) \citep{BrownMRSKDNSSAA20}, the issue of semantic inconsistency remains \citep{gan2023sensitivity}. \citet{rabinovich2023predicting} developed a benchmark dataset of high-quality paraphrases specifically for factual questions, serving as a testbed for evaluating semantic consistency in a QA context. Existing methods mainly addressed this issue through prompt engineering and data-driven SFT. For example, \citet{raj2023semantic} proposed an Ask-to-Choose (A2C) prompting method that can enhance both accuracy and semantic consistency in LLMs. \citet{zhou-etal-2022-prompt} used an unsupervised finetuning method. They took advantage of the fact that multiple prompts can be used to specify a single task and proposed to regularize prompt consistency, encouraging consistent predictions across this diverse set of prompts. Compared to previous methods, we use a model editing method to modify the output of specific model components in an LLM. This method is both transparent and computationally lightweight.

\noindent \textbf{Model Editing.}
The goal of model editing is to modify specific knowledge or control model behaviors without affecting the model’s performance on other tasks \citep{yao-etal-2023-editing}. There are mainly three types of editing methods: external memory-based methods, constrained fine-tuning methods, and locate-then-edit methods. 

Among them, (1) External memory-based methods introduce new parameters to update knowledge or change model behavior. An example is SERAC \citep{mitchell2022memory}, which used edit memory to store updated knowledge and a classifier to route between the edit memory and a pre-trained model.  (2) Constrained fine-tuning methods typically involve specific fine-tuning restrictions to regulate parameter updates, thus maintaining the model’s performance on unedited knowledge. For example, the method proposed by \citet{abs-2012-00363} implemented explicit constraints on model parameters to minimize the interference on the unmodified facts. (3) Locate-then-edit methods first identify relevant model parameters that store knowledge or steer model behavior, and then edit these parameters to achieve desirable outputs. \citet{meng2022locating} used causal analysis to find that factual knowledge is mainly stored in the intermediate MLP layer weights and subsequently used rank-one editing to modify model weights related to factual knowledge. \citet{li2023inference} demonstrated that by identifying specific attention heads and editing their activations, the likelihood of the model producing truthful output can be significantly enhanced. We adopt the ``locate-then-edit'' paradigm, motivated by the objective to improve the semantic consistency of the model while gaining insights into the components in an LLM contributing to this consistency.

\section{Preliminary}
\noindent \textbf{LLM representation.} The currently prevalent LLM adopts the decoder-only architecture. According to \citet{elhage2021mathematical}, an LLM mainly consists of three parts: token embedding, a sequence of decoder blocks, and token unembedding. Among them, token embedding is the process of mapping a token index to an embedding vector, while token unembedding is the reverse operation that maps the embedding back to the probability space of tokens, and then samples to obtain the index of the next token. The vast majority of parameters in LLM are composed of stacked decoder blocks, with each decoder block consisting of the components of multi-head attention and MLP, which can be represented by:

\begin{align}
\small
& a_{i}  = x_{i} + \sum_{j =1,...,J}h_{i,j} \label{eq_2} \\
& x_{i+1} = a_{i} + m_{i}, 
\label{eq_3}
\end{align}
where $x_i$ is the $i$-th decoder layer hidden states.  $h_{i,j}$ is the hidden output of the $j$-th attention head in the $i$-th layer.  $a_i$ is the residual output after multi-head attention. $m_i$ is the $i$-th MLP layer output. $x_{i+1}$ is the hidden output of the $i$-th decoder block and also the input of the $i\!+\!1$-th decoder block. 

\section{Methodology}
We use GPT-4 to construct prompt pairs that have the same semantics, and the target LLM outputs of these prompt pairs should ideally be consistent. However, when we use the target LLM to predict these prompt pairs, we obtain both consistent and inconsistent results. These inconsistent results are errors made by the target LLM. To locate the sources of these errors, we assume that if the model components (i.e. attention heads) behave similarly to the target LLM, then these components are actually the causes of the semantic consistency problems in the target LLM. On the other hand, those components that have large behavioral differences from the target LLM indicate that they are less relevant to the semantic consistency problems of the prompt pairs. Based on this assumption, we use the linear probing technique \citep{alain2016understanding} to identify the relevant components.

Next, we add semantic consistency biases to the identified components to correct their erroneous behavior. These biases are obtained by calculating the difference between the mass mean of the consistency samples and the mass mean of all samples on the corresponding components. These biases will shift the original activations of these components toward more semantically consistent directions.

As shown in Figure \ref{fig:my_image}, our method mainly consists of three steps, which are consistent prompt pairs construction, key components locating, and consistent-aware editing respectively. We will provide detailed explanations of these steps in the following sections.

\subsection{Consistent Prompt Pairs Construction}
We need to construct consistent prompt pairs for locating and editing an LLM. Specifically, we first construct a prompt pair set $\mathcal{D}$, whose element is represented as $[p,q]$. Here, $p$ represents the input prompt, and $q$ is the rephrased version of $p$, which can be generated using existing large-scale models like GPT-4.

Based on $\mathcal{D}$, we need a consistency evaluation label $c$ from the target LLM for key components locating and editing.  So we augment the consistency evaluation label $c$ to $[p, q]$ forming $([p,q],c)$.  In the case of NLU tasks, we determine consistency labels based on whether the predicted results are the same. For NLG tasks, we can utilize GPT-4 to assess the consistency. Subsequently, we add $c$ to each prompt pair in $\mathcal{D}$, obtaining the set $\mathcal{D'}$

\subsection{Key Components Locating}
With the constructed prompt pairs, we apply the linear probing method \citep{alain2016understanding} to identify which components have similar behavior to the LLM that determine the prompts' semantic consistency. Specifically, we divide the dataset $\mathcal{D'}$ into probe set $\mathcal{D'}_{probe}$ and locate set $\mathcal{D'}_{locate}$ following a 4:1 ratio. 

For each component, either an attention head or an MLP in any layer, we train a classifier that takes the concatenated hidden states as input and uses the consistency label $c$ as the ground truth label. These hidden states are the output hidden states of the component with respect to $p$ and $q$. The training data for this classifier comes from $\mathcal{D'}_{probe}$, and the testing data for this classifier comes from $\mathcal{D'}_{locate}$.

If the classifier achieves a high score on the locate set $\mathcal{D'}_{locate}$, it implies that the component and the overall LLM behave very similarly. On the other hand, a low score indicates that this component is less important for semantic consistency problems. We locate the top K components by ordering the classification accuracy. 

More specifically, given a sample $([p,q],c)$,  the linear classifier training feature for candidate MLP  and attention head are 
$f(m_{i}, p, q)$ and $f(h_{i,j}, p, q)$, respectively. 
\begin{align}
& f(m_{i}, p, q) = [m_{i}^{p^{last}}; m_{i}^{q^{last}}], \label{feat_m}\\
& f(h_{i,j}, p, q) = [h_{i,j}^{p^{last}};\ h_{i,j}^{q^{last}})], \label{feat_h}
\end{align}
where $p^{last}$ and $q^{last}$ indicates the last token of $p$ and $q$, and $m_{i}^{p^{last}}$ is the hidden output of the MLP layer in the $i$-th decoder block and $h_{i,j}^{p^{last}}$ is the hidden output of the $j$-th attention head in the $i$-th decoder block, all correspond to the last token. The reason why we only use the last token of $p$ and $q$ is that for a decoder-only architecture, the last token has visibility over all preceding tokens. Therefore, the hidden states corresponding to the last token can be considered a summary representation of the entire prompt. In this manner, we can construct training sets  $\mathcal{S}(m_{i})$ and $\mathcal{S}(h_{i,j})$
for training linear classifiers for $m_{i}$ and $h_{i,j}$, respectively. 
\begin{align}
    &\mathcal{S}(m_{i})\!=\!\{f(m_{i}, p, q), c \}_{([p,q],c)\in \mathcal{D'}_{probe}} \\
    & \mathcal{S}(h_{i,j})\!=\! \{f(h_{i,j}, p, q), c \}_{([p,q],c)\in \mathcal{D'}_{probe}}, 
\end{align}
where $\mathcal{S}(m_{i})$ and $\mathcal{S}(h_{i,j})$ are mapped from $\mathcal{D'}_{probe}$.

We train linear classifiers with $\mathcal{S}(m_{i})$ or $\mathcal{S}(h_{i,j})$, and then evaluate
these classifiers on $\mathcal{D'}_{locate}$. The top K components in LLM with the highest classification accuracy are used for model editing, as these components strongly affect the prompts' semantic consistency.

\subsection{Consistent-aware Model Editing}
Inspired by the work from \citet{li2023inference} and \citet{jorgensen2023improving}, we make specific adjustments to the hidden states of the top-K components, aligning their hidden states toward greater semantic consistency. 

Specifically, we add biases to these components, and the biases are obtained by calculating the difference between the mass mean of the consistency samples and the mass mean of all samples on the corresponding components. All of these samples are from $\mathcal{D'}_{probe}$. Formally, the biases for the candidate MLP and attention head are calculated by:

\begin{equation}
\begin{split}\small
    &b(m_{i})\!=\!\sum_{p,c=1} \frac{m_{i}^{p^{last}}}{N}\!-\!\sum_{p}\frac{ m_{i}^{p^{last}}}{M}, \\
    &b(h_{i,j})\!=\!\sum_{p,c=1}\frac{ h_{i,j}^{p^{last}}}{N}\!-\!\sum_{p}\frac{ h_{i,j}^{p^{last}}}{M},
\end{split}
\end{equation}
where $N$ is the number of the prompts in $\mathcal{D'}_{probe}$ with $c=1$, and $M$ is the number of all the instances in $\mathcal{D'}_{probe}$. After that, these biases are added to the hidden states of the selected Top-$K$ components, obtaining
$\hat{m}_{i}$ and $\hat{h}_{i,j}$ for the $K$ selected components.
\begin{align}
    & \hat{m}_{i} = m_{i} +\alpha \cdot b(m_{i}) \\
    & \hat{h}_{i,j} = h_{i,j} + \alpha \cdot  b(h_{i,j}).
\end{align}
Here, $\alpha$ is the hyperparameter that adjusts the strength of the activations shift. In our following experiments, we set its value to $5.0$. 

\section{NLU Benchmark Construction}
Currently, there are some NLG benchmarks related to the semantic consistency of LLM \citep{rabinovich2023predicting}. However, there is a relative scarcity of NLU benchmarks specifically designed for semantic consistency research. To address this gap, we propose a benchmark dataset for evaluating semantic consistency in NLU tasks. This benchmark comprises RobustSST2, RobustMRPC, and RobustBOOLQ, which are derived from the sentiment classification dataset SST2 \citep{socher2013recursive}, the text similarity dataset MRPC \citep{dolan2005automatically}, and the yes/no question-answering dataset BOOLQ \citep{clark2019boolq}, respectively. Our primary objective is to assess the semantic consistency of LLMs under synonymous task instructions for these datasets.

More specifically, we first generate 30  synonymous task instructions for each task dataset. For example, we feed the following prompt (bold font) to GPT-4 to generate the synonymous task instructions used for RobustSST2. 
\begin{table}[H] 
    \centering
    \small
    \begin{tabular}{p{0.9\linewidth}}
      \toprule
      \textbf{Rephrase the following sentence in 30 ways, while retaining the same meaning.} \\
      \texttt{Measure the polarity of this sentence and respond with either 'positive' or 'negative', give me one word.} \\
      \bottomrule
    \end{tabular}
\end{table}

Then, we slice the generated task instructions according to an 8:2 ratio, meaning the training set uses 24 instructions, while the test set uses 6 instructions.

For the training set, we constructed 24 synonymous prompts for each training instance, \emph{i.e.}, ${\text{prompt}_{i} = [\text{instruction}_{i}, \text{instance}_{train}]}_{i=1}^{24}$, and each prompt$_i$ has a label answer$_i$, which is consistent across these 24 prompts.

Additionally, to create consistent prompt pairs for model editing, we generated $C_{24}^{2}$ prompt pairs by iterating through all possible combinations of these 24 prompts for each training instance. Then, we utilized the target LLM to assess whether the predictions generated for each prompt pair were consistent and obtain the relevant consistency evaluation label $c$. Lastly, we sampled 250 instances from both the $c=0$ and $c=1$ categories, yielding a total of 500 instances used for model editing.

The instance in the test set is different from the instance in the training set. Each instance in the test set includes a constructed prompt along with its corresponding answer $[\text{prompt}_{test}, \text{answer}_{test}]$. In specific, we first use the left 6 task instruction to construct relevant prompts \emph{i.e.}, $\{\text{prompt}_{i} = [\text{instruction}_{i}, \text{instance}_{test}]\}_{i=25}^{30}$. Subsequently, We perform sample selection for these 6 prompts to construct a test instance. The selection rule is that if these 6 prompts yield the same result, we randomly choose 1 prompt to construct the sample. However, if they predict $N$ different outcomes, we select 1 prompt from each of the $N$ distinct results from N samples. By employing this approach, we can select hard negative samples while retaining examples that the LLM could originally predict correctly, thereby enhancing the diversity of our test data.

\section{Experiments}
\subsection{Datasets}
To verify the effectiveness of the model editing method for addressing the issue of prompt semantic consistency, we conducted tests on both NLU and NLG tasks.
For NLU evaluation, we utilized the specially constructed RobustMRPC, RobustSST2, and RobustBOOLQ datasets. For NLG evaluation, we selected the sport and capital categories from the PopQA question-answering dataset as described by  \citet{rabinovich2023predicting}. Detailed data statistics are shown in Table \ref{tab:data_stat}.

\begin{table}[h]\footnotesize
\centering
\begin{tabular}{@{}lcc@{}@{}}
\toprule
\textbf{Task Category}  & \textbf{Datasets} & \textbf{Number of test cases} \\ \midrule
NLU & RobustMRPC & 408 \\ 
NLU & RobustSST2 & 872 \\ 
NLU & RobustBOOLQ & 1000 \\ 
\midrule
NLG & PopQA\_sport & 3829 \\
NLG & PopQA\_capital & 4515 \\
\bottomrule
\end{tabular}
\caption{Data statistics of Evaluation Datasets.}
\label{tab:data_stat}
\end{table}

\subsection{Evaluation Metrics}
For NLU tasks, we evaluate its task performance by testing the overall accuracy of classification results across different instruction templates. To assess semantic consistency, we measure the standard deviation of the accuracy across these various instruction templates.
For the NLG task, accuracy and mean pairwise cosine similarity metrics introduced by \citet{rabinovich2023predicting} are employed to evaluate the model's task performance and its semantic consistency respectively. It is worth noting that lower standard deviation and higher mean pairwise cosine similarity are both indicative of better semantic consistency.

\subsection{Key Components Locating Result}
We utilize the LLama2-7B chat-version model \citep{touvron2023llama} as the target LLM to analyze the impact of candidate model components, such as attention heads and MLPs, on LLM’s semantic consistency problem. Next, we visualize the locating results of these components, with brighter squares (i.e. yellow squares) highlighting areas of high locating accuracy, indicative of a strong correlation with semantic consistency.

\begin{figure}[ht]
    \centering
    \begin{subfigure}{\columnwidth}
        \includegraphics[width=\columnwidth]{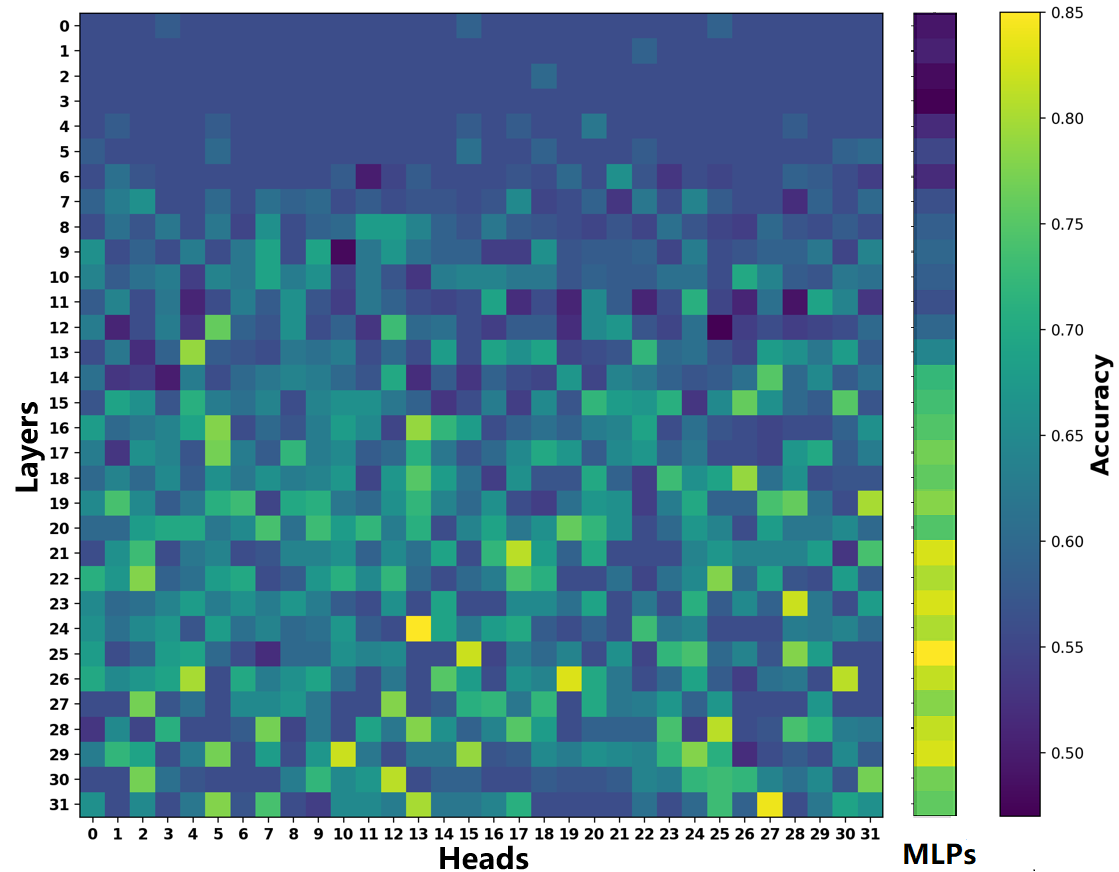}
        \caption{Visualization Result on RobustSST2.}
    \end{subfigure}
    \begin{subfigure}{\columnwidth}
        \includegraphics[width=\columnwidth]{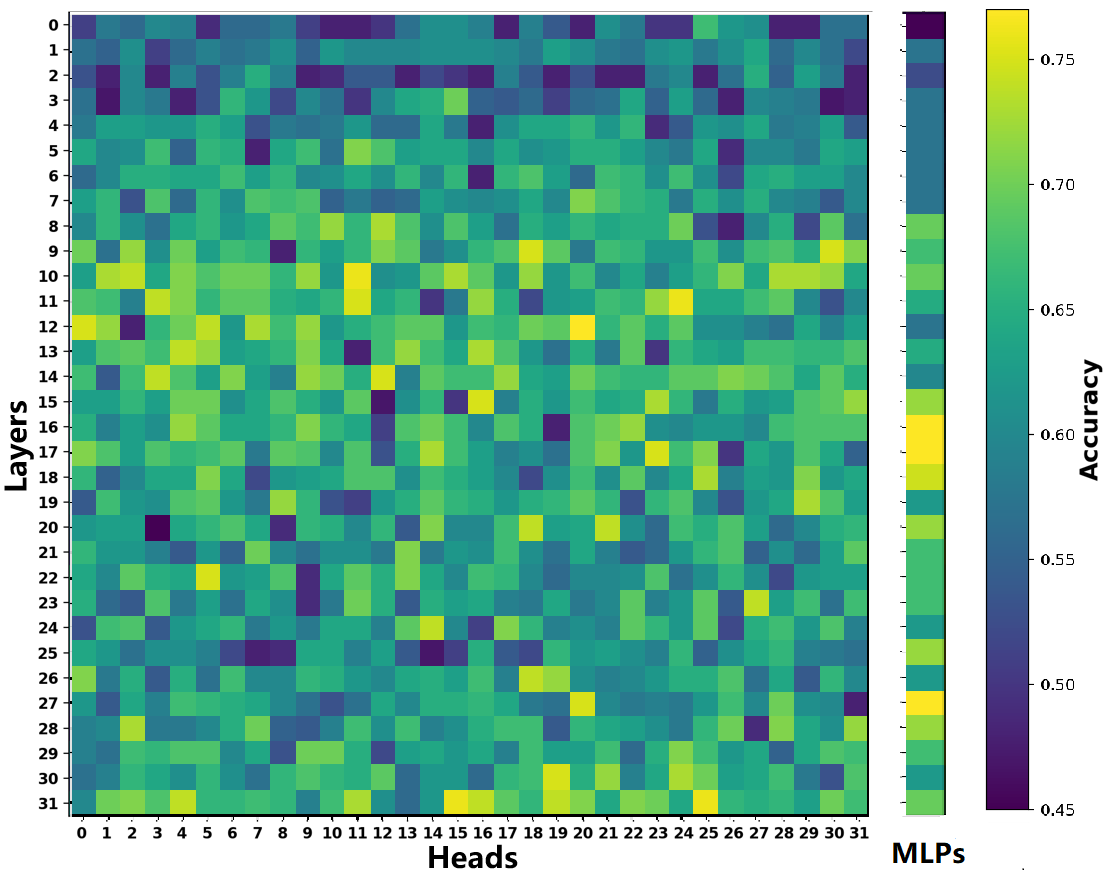}
        \caption{Visualization Result on PopQA\_capital.}
    \end{subfigure}
    \caption{The visualization experiments on the RobustSST2 (NLU) and PopQA\_capital (NLG) dataset. The horizontal axis represents the attention heads and the MLP in certain layer, while the vertical axis indicates the layer number. The column on the right shows the locating accuracy of attention heads or the MLPs. Brighter Squares indicate high locating accuracy.}
    \label{fig:my_label}
\end{figure}

As the visualization result shown in Figure \ref{fig:my_label}, we find that there exists a notable concentration of yellow squares between layers 11 and 32, suggesting that attention heads and MLPs in the mid to final LLM's decoder blocks are highly relevant to semantic consistency. 

Furthermore, our findings suggest that model components in the initial layers of transformer blocks exert negligible influence on semantic consistency. Their locating accuracy for synonymous samples hovers around 50\%, equivalent to random chance, indicating these samples are treated as identical by these components. This distinction underscores the nuanced role of the model components across different layers in influencing LLM's semantic consistency.

\subsection{Model Editing Experimental Result}

\begin{table}[htbp]\footnotesize
\centering
\begin{tabular}{@{}l@{  }c@{  }c@{  }c@{}}
\toprule
\textbf{Method}  & \textbf{RobustMRPC} & \textbf{RobustSST2}  & \textbf{RobustBOOLQ}   \\ \midrule
LLama2-7B & $67.15_{ \pm 5.36}$ & $85.66_{ \pm 4.88}$ & $46.40_{ \pm 10.55}$  \\
+Editing  & $68.62_{ \pm 4.47}$ & $89.90_{ \pm 4.54}$ & $57.50_{ \pm 5.10}$ \\ \bottomrule
\end{tabular}
\caption{
Main experiment result is on NLU datasets. The notation $67.15_{\pm5.36}$ indicates an average test set accuracy of $67.15$ with a standard deviation of $\pm5.36$.} 
\label{tab:acc_nlu}
\end{table}

\begin{table}[htbp]\footnotesize
\centering
\begin{tabular}{@{}lcc@{}}
\toprule
\textbf{Method} & \textbf{PopQA\_sport}  & \textbf{PopQA\_capital} \\ \midrule
LLama2-7B  & $50.83_{/0.79}$ & $73.33_{/0.73}$ \\
+Editing & $53.20_{/0.80}$ & $74.36_{/0.77}$ \\ \bottomrule
\end{tabular}
\caption{Main experiment on NLG Tasks. 
The notation $50.83_{/0.79}$ indicates an average test set accuracy of $50.83$ with a mean pairwise cosine similarity of $0.79$.} 
\label{tab:acc_nlg}
\end{table}

% We evaluate the efficacy of employing semantically consistent prompt pairs for model editing. 
Our comprehensive analysis, as presented in Table \ref{tab:acc_nlu}, and Table \ref{tab:acc_nlg}, demonstrates that our editing method can significantly enhance both semantic consistency and task performance across a variety of NLU and NLG tasks. Specifically, we observed notable reductions in the standard deviation for semantic consistency assessments on the RobustMRPC, RobustSST2, and RobustBOOLQ datasets, with decreases of 0.89, 0.34, and 5.45, respectively. Moreover, the accuracy of model performance experienced substantial improvements, showing increases of 1.47\%, 4.24\%, and 11.1\% across these datasets, respectively. For NLG tasks, we noted improvements in semantic consistency score by 1.0\% and 4.0\%, respectively, while the accuracy in these tasks rose by 2.37\% and 1.03\%, respectively.

The findings from these experiments clearly support the conclusion that adjustments to the outputs of the top-K model components can significantly enhance the model's semantic consistency and task performance. Importantly, these advancements are achieved without the need for altering the model's underlying parameters.

\subsection{Ablation Study}
\subsubsection{The Influence of Hyperparameter K}
\begin{figure}[htbp]
    \centering
     \includegraphics[width=\columnwidth]{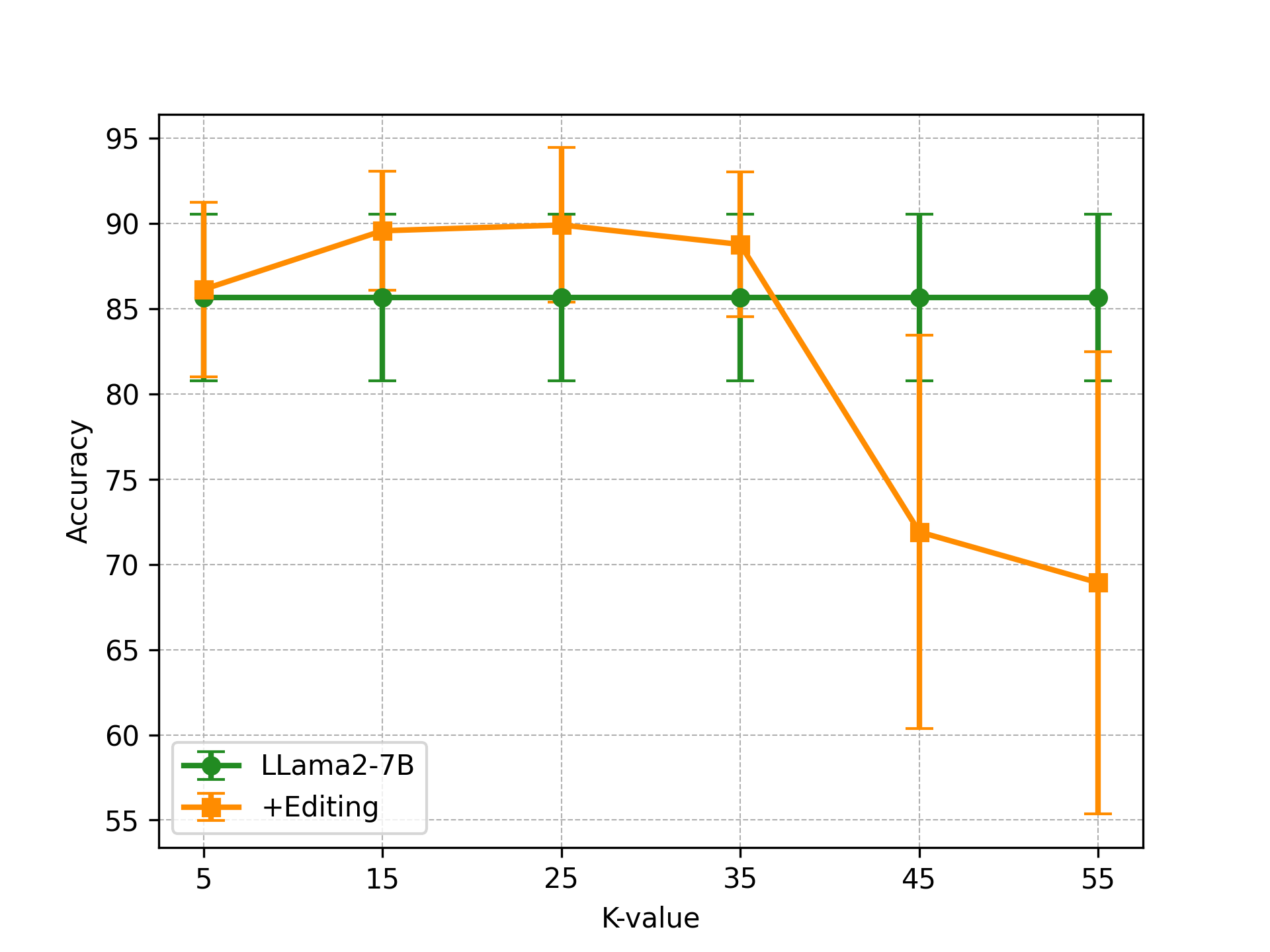}
    \caption{The performance of the proposed model editing method with different $K$-values. }
    \label{fig:abl_params}
\end{figure}

We investigate the impact of the $K$-value on the experimental setting, selecting the RobustSST2 dataset for our analysis, with the $k\in\{5, 15, 25, 35, 45, 55\}$. As demonstrated in Figure \ref{fig:abl_params}, it is observed that the edited model achieves the highest accuracy when the $K$-value is equal to 25. Conversely, when the $K$-value is equal to or greater than 35, a decline in model accuracy is noted. The experimental findings underscore the critical importance of selecting an appropriate number of editing heads. It is evident that excessive model editing can result in the collapse of the model.

\subsubsection{The Influence of Hyperparameter $\alpha$}
We also perform an ablation study to investigate the impact of the activations shift strength hyperparameter $\alpha$ on the LLM's task performance and semantic consistency. Specifically, we evaluate the LLM edited using the proposed method on RobustSST2 dataset with $\alpha \in \{1.0, 3.0, 5.0, 7.0, 9.0\}$, respectively.

As depicted in Table \ref{tab:abl_alpha}, the performance of the edited LLM exhibits robustness within a specific range of $\alpha$ values (1.0 to 7.0). Across this range, our methods consistently enhances both the task performance and semantic consistency of the LLM. Concurrently, it is anticipated that an excessively large activations shift strength ($\alpha=9.0$) can lead to a decline in model performance. It is noted that at $\alpha=3.0$, the LLM achieves better results than those reported in Table \ref{tab:acc_nlu}, highlighting the need to optimize the value of $\alpha$ in practice.  

%This is because we did not intentionally optimize the $\alpha$ value and set it to $5.0$ in all experiments, which also highlights the potential of our proposed method. -- 这个感觉不是原因， 最好不写

\begin{table}[h]\footnotesize
\centering
\begin{tabular}{@{}lcc@{}}
\toprule
\textbf{Method}  & \textbf{RobustSST2}  \\ \midrule
LLama2-7B & $85.66_{ \pm 4.88}$ \\ 
+Editing ($\alpha$=1.0) 	 & $89.90_{ \pm 4.54}$ \\
+Editing ($\alpha$=3.0) 	 & $89.90_{ \pm 2.98}$ \\
+Editing ($\alpha$=5.0) 	 & $89.90_{ \pm 4.54}$ \\
+Editing ($\alpha$=7.0) 	 & $88.53_{ \pm 5.09}$ \\
+Editing ($\alpha$=9.0) 	 & $83.48_{ \pm 7.50}$ \\ 
\bottomrule
\end{tabular}
\caption{The performance of the proposed model editing method under different $\alpha$ values.} 
\label{tab:abl_alpha}
\end{table}

\subsubsection{The Influence of the Model Components Selecting Strategy and Editing Direction}
We design two ablation studies to investigate effects of our editing method on the selection strategy of model components and the direction of editing. In the first study, we randomly select a number of model components equivalent to those obtained in our editing method, with the goal of evaluating the effectiveness of our components selection strategy on the performance of the model. The second study involves altering the editing directions to random directions based on biases randomly generated from a normal distribution.

\begin{table}[h]\footnotesize
\centering
\begin{tabular}{@{}lcc@{}}
\toprule
\textbf{Method}  & \textbf{RobustMRPC}  \\ \midrule
LLama2-7B & $67.15_{ \pm 5.36}$ \\ 
+Editing & $68.62_{ \pm 4.47}$ \\ 
w/ random components & $61.51_{ \pm 10.86}$ \\
w/ random direction & $64.46_{ \pm 6.20}$ \\ 
\bottomrule
\end{tabular}
\caption{Ablation studies for the influence of model components selecting strategy and the editing direction. The term ``random components'' refers to the strategy of randomly selecting the same number of components as in our editing method. The ``random direction'' is the approach of randomly selecting editing directions. } 
\label{tab:abl_head}
\end{table}

Table \ref{tab:abl_head} demonstrates that on the RobustMRPC dataset, both semantic consistency and the task performance of the model suffer when random model components selection or random direction editing are applied. Compared to the unedited LLama2-7B chat-version model, semantic consistency experiences a decline of 5.5 and 0.84 points (measured in the standard deviation of accuracy), while accuracy drops by 5.64\% and 2.69\%, respectively. The findings highlight the critical role of model components selection and editing direction decision in applying model editing to enhance the semantic consistency and task performance of an LLM.

\subsubsection{Out-of-domain Experiment Result}
We evaluate the performance of the edited LLama2-7B chat-version model on out-of-domain datasets. Specifically, after editing the model on the MRPC dataset, we test its performance on four OOD datasets: AG News for news categorization \citep{Zhang2015CharacterlevelCN}, IMDB for movie reviews sentiment classification  \citep{imdb_data}, and both CNN/Daily Mail \citep{see-etal-2017-get} and XSum \citep{Narayan2018DontGM} for news summarization, drawing a sample of 500 instances from each for evaluation. For AG News and IMDB, accuracy serves as the evaluation metric, while for CNN/Daily Mail and XSum, we apply the ROUGE-L metric \citep{lin2004rouge} for assessment. 

\begin{table}[ht]\footnotesize
\centering
\begin{tabular}{@{}lcc@{}}
\toprule
\textbf{Model}  & \textbf{AG\ News} & \textbf{IMDB}  \\ \midrule
LLama2-7B & 70.00 & 88.60 \\
+Editing & 70.20 & 89.40 \\ \bottomrule
\end{tabular}
\caption{Evaluation of OOD performance on AG News and IMDB datasets using a subset of 500 instances from each.}
\label{tab:ood_nlu}
\end{table}

\begin{table}[ht]\footnotesize
\centering
\begin{tabular}{@{}lcc@{}}
\toprule
\textbf{Model}  & \textbf{CNN/Daily Mail} & \textbf{XSum}  \\ \midrule
LLama2-7B & 21.36 & 14.28 \\
+Editing  & 21.14 & 14.45 \\ \bottomrule
\end{tabular}
\caption{Experiment results on OOD performance with 500 instances from CNN/Daily Mail and XSum.}
\label{tab:ood_nlg}
\end{table}

It can observed in Tables \ref{tab:ood_nlu} and \ref {tab:ood_nlg} that the model's performance remains consistent across most datasets, demonstrating the potential generalization capability of the proposed editing method across OOD tasks.

\subsubsection{Comparison with the SFT Method}
To compare our method with the STF approach, we also employ the LLama2-7B-Chat model as the base model. Specifically, we generate additional training samples by paraphrasing the task prompt with the same semantic meaning as the fine-tuning data. The number of training samples is identical to that of our model editing method (500 for each relevant task).

\begin{table}[h]\footnotesize
\centering
\begin{tabular}{@{}l@{  }c@{  }c@{  }c@{}}
\toprule
\textbf{Method}  & \textbf{RobustMRPC} & \textbf{RobustSST2}  & \textbf{RobustBOOLQ}   \\ \midrule
LLama2-7B & $67.15_{ \pm 5.36}$ & $85.66_{ \pm 4.88}$ & $46.40_{ \pm 10.55}$  \\
+Editing & $68.62_{ \pm 4.47}$ & $89.90_{ \pm 4.54}$ & $57.50_{ \pm 5.10}$ \\
% L2+UPEGood & $68.62_{ \pm 4.89}$ & $90.02_{ \pm 4.40}$ & $64.80_{ \pm 8.14}$ \\
+SFT & $80.14_{ \pm 2.40}$ & $91.39_{ \pm 1.94}$ & $81.80_{ \pm 3.87}$ \\ \bottomrule
\end{tabular}
\caption{Comparison of performance and consistency between the SFT and our method on NLU datasets.}
\label{tab:sft_nlu}
\end{table}

\begin{table}[h]\footnotesize
\centering
\begin{tabular}{@{}lcc@{}}
\toprule
\textbf{Method} & \textbf{PopQA\_sport}  & \textbf{PopQA\_capital} \\ \midrule
LLama2-7B  & $50.83_{/0.79}$ & $73.33_{/0.73}$ \\
+Editing & $53.20_{/0.80}$ & $74.36_{/0.77}$ \\
% L2+UPEGood & $45.86_{ \pm 0.75}$ & $73.74_{ \pm 0.75}$ \\
+SFT & $74.03_{/0.95}$ & $70.89_{/0.91}$ \\ \bottomrule
\end{tabular}
\caption{Comparison of the performance and consistency between the SFT and our method on NLG datasets.}
\label{tab:sft_nlg}
\end{table}

The experimental results presented in Tables \ref{tab:sft_nlu} and \ref{tab:sft_nlg} suggest that while our method enhances both semantic consistency and task performance, the magnitude of improvement is not as pronounced as that achieved by SFT. Notably, SFT outperforms our editing method on RobustMRPC, RobustSST2, RobustBOOLQ, and PopQA\_{sport}. The only exception is the NLG task’s PopQA\_{capital} dataset, where our method slightly surpasses SFT (74.36 vs. 70.89). SFT achieves superior performance by precisely adjusting model parameters based on backpropagation (BP). In contrast, our editing method prioritizes model components interpretability, adjusting the output of the key components instead. There is ample space for optimization. 

However, in terms of computational resource consumption, our method exhibits a significant advantage over SFT, as shown in Table \ref{tab:gpu_hour}. For instance, SFT requires 2.02 GPU hours to complete the task on the RobustSST2 dataset, whereas our method achieves the same task with a significantly reduced computational resource (0.11 GPU hours).

\begin{table}[h]\footnotesize
\centering
\begin{tabular}{@{}lccc@{}}
\toprule
\textbf{Method} & \textbf{+SFT}  & \textbf{+Editing} & \textbf{Efficiency} \\ \midrule
 RobustSST2 &  2.02 &  0.11  & 18X \\
 RobustMRPC &  2.80   & 0.12 & 23X   \\
 RobustBOOLQ & 1.68  & 0.14 & 12X    \\
 PopQA\_sport  & 1.87 & 0.10 & 18X \\
 PopQA\_captial & 1.93 & 0.10 & 19X \\ \bottomrule
\end{tabular}
\caption{The comparison of the computational cost between SFT and our method in terms of GPU hour.}
\label{tab:gpu_hour}
\end{table}
Furthermore, we conduct an experiment to investigate the pontential complementary effect between SFT and our proposed model editing approach. Specifically, we apply SFT to the LLM followed by the proposed model editing to the fine-tuned model. As shown in Table \ref{tab:sft_post_edit} (``+SFT\&Editing''), our preliminary experimental results on RobustSST2 and RobustBOOLQ datasets indicate that the task performance and semantic consistency of an LLM can be further enhanced by applying model editing techniques to a fine-tuned model. While the magnitude of improvement may not be substantial, it nonetheless demonstrates the potential of a two-stage approach. We intend to conduct further investigation of this approach in future work.

\begin{table}[ht]\footnotesize
\centering
\begin{tabular}{@{}l@{  }c@{  }c@{  }c@{}}
\toprule
\textbf{Method} & \textbf{RobustSST2}  & \textbf{RobustBOOLQ}   \\ \midrule
LLama2-7B & $85.66_{ \pm 4.88}$ & $46.40_{ \pm 10.55}$    \\
+Editing & $89.90_{ \pm 4.54}$ & $57.50_{ \pm 5.10}$   \\
+SFT &  $91.39_{ \pm 1.94}$ & $81.80_{ \pm 3.87}$   \\ 
+SFT\&Editing & $91.51_{ \pm 1.91}$ & $81.80_{ \pm 3.63}$ &  \\ \bottomrule
\end{tabular}
\caption{Comparison of performance and semantic consistency between the SFT, Editing and SFT\&Editing method on RobustSST2 and RobustBOOLQ datasets.}
\label{tab:sft_post_edit}
\end{table}

\section{Conclusion}
This paper presents the first analysis of the internal mechanism aspects of an LLM that contribute to the problem of semantic inconsistency. We can precisely diagnose the key components that contribute to a model's semantic consistency. Based on this finding, we propose a model editing method that directly injects biases into the model components of an LLM without mass-manipulating model parameters. The proposed method can significantly improve both semantic consistency and the performance of LLMs on the constructed NLU and open-source NLG datasets. Also, our methods exhibit promising generalization capabilities on four OOD task datasets.

\section*{Limitations}
Our study reveals that semantic consistency is correlated with both attention heads and MLPs in an LLM. However, attention heads tend to have a more predominant influence on an LLM than MLPs with the majority of editing operations focusing on them. Future research will focus on exploring the role of MLPs in the semantic consistency of LLMs.

Despite achieving comparable results on OOD settings, our editing method is not sufficiently validated in terms of other metrics, like locality and portability \citep{yao-etal-2023-editing}. Therefore, more rigorous and effective testing methods are required to evaluate the performance of the proposed method. 

We aim to develop an interpretability-oriented approach to enhance the semantic consistency of LLMs.
Despite our model editing method being comparably transparent and computationally efficient, it still lags behind an SFT approach in terms of performance. In the future, we plan to extend our research to identify the circuits \citep{elhage2021mathematical} related to semantic consistency and understand their causal mechanisms. In this way, we can further advance the development of effective techniques that improve the semantic consistency of LLMs while prioritizing interpretability and efficiency.

\bibliography{custom}
% \end{CJK*}
\end{document}